\documentclass{article}

    \usepackage[final]{neurips_2023}

\usepackage[utf8]{inputenc} 
\usepackage[T1]{fontenc}    
\usepackage{hyperref}       
\usepackage{url}            
\usepackage{booktabs}       
\usepackage{amsfonts}       
\usepackage{nicefrac}       
\usepackage{microtype}      
\usepackage{xcolor}         
\usepackage{mathtools}
\usepackage{tikz}
\usepackage{caption}
\usepackage{subcaption}

\title{LLMs grasp morality in concept.}

\author{%
  Mark Pock\thanks{Both authors contributed equally to this research.} \\
  University of Washington\\
  \texttt{markpock@uw.edu} \\
  \And
  Andre Ye$^*$ \\
  University of Washington \\
  \texttt{andreye@uw.edu} \\
  \And
  Jared Moore\\
  Stanford University\\
  \texttt{jlcmoore@stanford.edu}\\
}

\begin{document}

\newcommand{\jared}[1]{\textcolor{blue}{[jared: #1]}}
\newcommand{\andre}[1]{\textcolor{red}{[andre: #1]}}
\newcommand{\markC}[1]{\textcolor{purple}{[mark: #1]}}


\maketitle




\begin{abstract}

Work in AI ethics and fairness has made much progress in regulating LLMs to reflect certain values, such as fairness, truth, and diversity.
However, it has taken the problem of \textit{how LLMs might `mean' anything at all} for granted.
Without addressing this, it is not clear what imbuing LLMs with such values even \textit{means}.
In response, we provide a \textit{general} theory of meaning that extends beyond humans. 
We use this theory to explicate the precise nature of LLMs as meaning-agents. 
We suggest that the LLM, by virtue of its position as a meaning-agent, already grasps the constructions of human society (e.g. morality, gender, and race) \textit{in concept}.
Consequently, under certain ethical frameworks, currently popular methods for model alignment are limited at best and counterproductive at worst.
Moreover, unaligned models may help us better develop our moral and social philosophy.




\end{abstract}

\maketitle


\section{Introduction}\label{sec:intro}


The advent of increasingly powerful LLMs poses many important ethical problems, 
such as the reproduction of harmful social biases~\citep{Mehrabi2019ASO} or limited viewpoints~\citep{Santy2023NLPositionalityCD},
hallucination of false information~\citep{Ji2022SurveyOH},
and producing morally or socially objectionable outputs~\citep{Choi2022KnowledgeIP}. 
In setting out to solve these problems, AI ethics and AI fairness/bias research 
have often asked questions and provided solutions that are deeply empirical or pragmatic in nature.
These inquiries are often motivated by intuitions about human values and human-AI relationships.
Relying on these intuitions tends to reproduce uncritical, anthropocentric accounts of AI, limiting our understanding of AI as independent systems.
For instance, consider a hypothetical model which, like Delphi \citep{Jiang:Delphi}, produces moral judgements based on text inputs. Suppose it marks the objectionable statement ``helping a friend spread fake news'' as morally unproblematic.
Rather than immediately identifying this as moral error and modifying the model to conform to our existing judgements, we might read the prediction as indicating a socially real system of moral meaning.



As work in AI ethics and fairness grows, we ought to explore these foundational questions which have been overlooked in favor of empirical ones.
In particular:
What are the meanings of `values' and `morality'?
What and how does an LLM mean?
In what ways do LLMs and humans \textit{mean} -- including meaning values and morality -- differently? 


To address these questions, we set forth a \textit{general} theory of meaning in $\S$ \ref{sec:theory}.
It is general in two ways that previous theories of meaning are not.
First, it has explanatory power over both humans and models, as well as other systems.
Second, it aims to be multimodal and account for all meaning in experience and not just language.
We then move on in $\S$ \ref{sec:model} to discuss how LLMs can be interpreted through such a theory of meaning. 
In particular, we give a reading of a prototypical general-purpose LLM whose outputs admit moral judgements. Such an LLM is one that, like ChatGPT, Delphi, or Bard, exists in a public setting as a kind of question-answer machine. Crucially, this means that an LLM is defined as a function which accepts \textbf{text input}. In homage to Delphi, we call these LLMs \textbf{oracles}. Though all the models we mentioned are aligned chat models, we use oracle much more generally in a way that can also refer to unaligned base models.

Lastly, in $\S$ \ref{sec:moral}, we discuss the social possibilities of oracles and how oracles are already related to social meaning and human values. 
We understand values like fairness, justice, and liberty as ``social objects'' with complex and diverse genealogies. 
This analysis ultimately allows us to determine the relationship between models and values. Ultimately, we conclude that there is a very real sense in which unaligned LLMs are \textit{already aligned to human values} (in concept).

\section{A General Theory of Meaning}\label{sec:theory}

We present a general theory of meaning applicable across agent types (human, machine, etc.) and multimodal contexts. 
This theory does not aim to describe its own `implementation' in meaning-agents. 
Instead, this theory provides an efficacious model for discussing the many similarities between meaning-agents.
It is helpful to read this section as a genealogy of `social objects' such as values (fairness, liberty, equality) and social categories (race, class, gender).
Such social objects begin, spurred on by context, inside the individual as abstract and become actualized over time.
Eventually, they enter into the shared social-material world, altering what was already there.
These social constructions begin to have material effects, which become new contexts that ground concretization.
``Social kinds'' are one prominent example of this ``looping effect'' \citep{Hacking:LoopingEffects}.



\begin{table}[!ht]
\centering
\begin{tabular}{l|l}
\toprule
Term            & Brief Description \\
\midrule
Signification   & A relationship between two experiences, one of which picks out the other. \\
Context         & A material/social arrangement which imposes structure upon experience. \\
Sign            & An experience taken as signifying another experience. \\
Object          & An experience taken as being signified by another experience. \\
Concept         & An internalization of a context, regulating a set of sign-object relationships. \\
Determination   & The process of an object acquiring a signification. \\
Concretization  & A process of individual objects being determined by the context. \\
Inscription     & A process of altering the context through the exercise of concepts. \\
Social Totality  & The social arrangement governing a community of individuals. \\
\bottomrule
\end{tabular}
\end{table}

\subsection{Snapshots of meaning}\label{sec:theory:synchronics}

We will begin by discussing how our general theory of meaning operates at snapshots in time.
First, we start with pure individual experience in the snapshot before meanings emerge.
Importantly, according to our model of experience, all experiences begin by grasping a `thisness' of something \citep{Hegel:PhG}.
For example, suppose there is a table in a room.
The table might be brown, mottled with age, stained with food, etc.
This table has a variety of properties.
But an individual, upon first seeing the table, does not stop to pay attention to its brownness, its age, its uncleanliness, and so on.
Instead, this individual merely grasps that this table is a thing -- the `thisness' of the table.
If the individual has never seen a table before, they might not even grasp that it is in fact a table.
The individual's experience of the table is thus \textit{surprisingly empty}.
This experience has no internal contents or particularities, it just \textit{is}.
Because of this, we say that \textit{experiences begin as }\textbf{abstract}.
This initial abstractness of the experience allows us to avoid the pitfalls of qualia (ineffable, intrinsic, immediate, private \citep{Dennett:QuiningQualia} subjective experiences such as the redness of an apple).

Over time, experiences become interrelated; they fall into certain roles with respect to one another.
Now, we will consider a snapshot wherein meanings have begun to develop -- that is, experiences have become related.
We call a relation between experiences a \textbf{signification}.
With a signification, one experience picks out another experience for an agent.
We call the experience that does the picking out the \textbf{sign}.
The experience being picked out is the \textbf{object} of that sign.
This description applies to all relations between experiences for any agent.
For an animal, a loud noise signifies a potential threat.
For humans using language, a name signifies a person.
The way in which the loud noise signifies the threat is very different from the way in which a name signifies a person.
The former does so through an automatic physiological response, the latter does so through language.
This `way of signifying' is the result of an external arrangement, social or material, imposing itself on experience.
For the animal, this arrangement is its evolutionary biology.
For a human using language, this arrangement is the language-game (a set of conventions that regulate the usage of language in a certain context \citep{Wittgenstein:PhilosophicalInvestigations}). 
This external arrangement is the \textbf{context}.
We might implicitly internalize the context (perhaps by learning the rules of the language-game over time).
A way of internalizing a context with respect to a set of meanings is a \textbf{concept}.
Over time, concepts enable individuals to develop more significations.
Thus, the central maxim of our theory is ``The \textbf{concept} makes possible the \textbf{signification} of the \textbf{object} by the \textbf{sign}.''


\begin{figure}
    \centering
    \includegraphics[width=6cm]{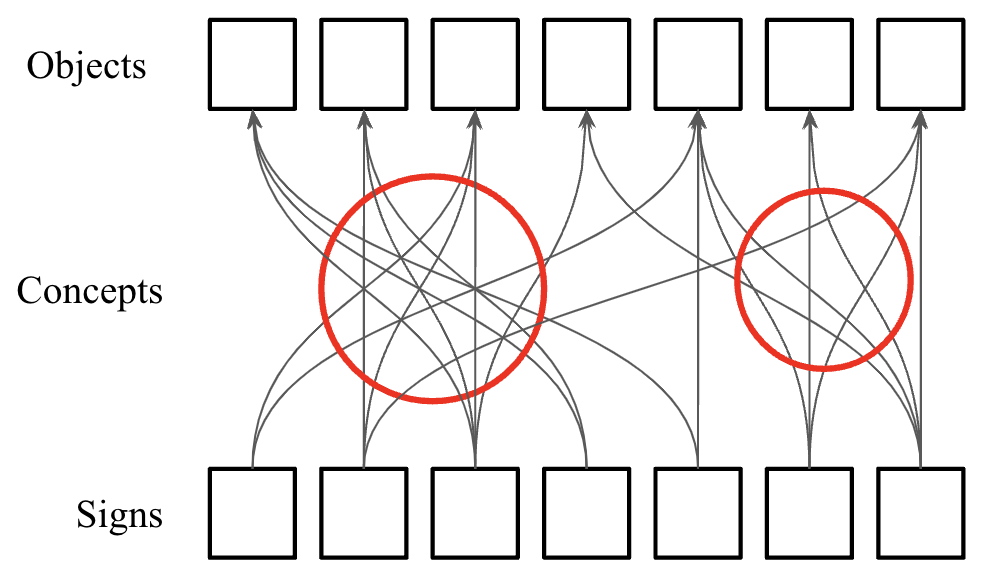}
    \caption{Signification relations between \textit{signs} and \textit{objects}. Both signs and objects  have no internal content; they are determined by their relations ($\S$ \ref{sec:theory:synchronics}). Groups of significations are called \textit{concepts}.}
    \label{fig:enter-label}
\end{figure}

Recall that objects and signs are both just experiences. This allows us to explain meaning using only two sorts: \textit{experiences} and \textit{the relationships between experiences} (significations).\footnote{This two-sortedness bears a certain similarity to category theory with its notions of objects and morphisms. This is no accident \citep{Tsuchiya:CategoryTheoryNeuroscience}.
}
This allows us to avoid complications with objects that do not exist in the material world.
Fictional characters are a classic example -- what does a name like ``Hamlet'' refer to? For our theory, it is the experiences signified by ``Hamlet.''
We also avoid the problem of whether proper names have descriptions or just tag their real objects by obviating the question of `real' referents entirely \citep{Kripke:NamingNecessity}.

\subsection{A genealogy of the object}\label{sec:theory:diachronics}
So far, our theory of meaning has provided the grounds for understanding how meaning operates at a single snapshot in time through four roles: the concept, the sign, the signification, and the object.
We now turn to the ways in which meaning applies itself over time to produce the object (i.e., its genealogy). 
Recall that we described experiences as initially abstract. 
Now, as experiences become objects of signs and signs for objects, they slowly become \textbf{concrete}. 
The original pure `thisness' is replaced by a rich concept, full of links to other experiences. 
Eventually, this concrete object enters the social-material world, thereby becoming fully concrete. 
These two processes are appropriately called \textbf{concretization} and \textbf{inscription}. 
Concretization is the process whereby an object becomes more concrete through acquiring significations. 
Each acquisition of a signification is called a \textbf{determination}.
More determined objects are thus more concrete.
Inscription is the process whereby individual meanings enter the social-material world. 
This is a process of realizing the object, actualizing what before only ever existed as an experience.

\paragraph{Concretization}
Our original usage of the term `concept' to mean an internalized context may seem strange. The term `concept' usually indicates an idea, something individuals have direct access to. Usually, this idea is thought of as having (or being) content. However, these two notions prove to be the same. First, we will understand how the colloquial notion appears in our theory. We do not discuss the \textit{internal} content of experiences, so experiences must instead acquire content by determination. This \textit{external} content is their concept. This comports with the colloquial sense of `concept': the concept of an experience is the things it signifies and the things that signify it. The concept of fairness is all things fairness means and all things that mean fairness. Fairness is what it is because it is (in part) impartiality and (in part) a standard by which to measure behavior (among other significations). The concept of fairness is described by an enumeration of its significations.

This definition of the concept (all the significations of an experience) aligns with our first definition of the concept (a way of internalizing a context).
Because a context is an external arrangement, it has a structure of `real' social or material objects and the relations between them \citep{Millikan:BeyondConcepts}. When these objects are experienced by individuals, they attempt to replicate this structure. The result of this replication is in fact the set of significations of experiences (our previous definition of concept).

Because concretization is a process of determination (acquiring significations), both the creation and adjustment of concepts are processes of concretization. Operationally, creation and adjustment differ heavily. Efficacious creation of a concept requires that the meaning-agent clearly prioritize a context. Once this prioritization is done, the actual creation follows as something simple. For example, the context ``the English language-game'' is prioritized
\footnote{There is a strong link between prioritization and intentionality via the notion of unitarity (see $\S$\ref{appendix:possibility}). 
}
; a word such as `table' is determined against an object, a `real' table (potentially) in the world, therefore becoming concretized.
However, a failure to adequately prioritize a context can result in scattered, non-efficacious concepts.
Adjustment, however, occurs under a predetermined context. If `table' is made to refer to all tables instead of just a singular table in the world, the original determination continues to exist under a different context.

\paragraph{Inscription}
Broadly speaking, inscription is the process of writing objects into the world -- creating or altering social material objects that correspond to experiential objects. \textit{However, inscription is not as simple as a direct transfer from the experiential world to the social-material world}. This may seem problematic -- after all, communication depends entirely on inscription. However, since communication is possible, we can infer that inscription must have certain properties.


For meaning to be possible, we posit that meaning can in fact function when the referents of terms in a language are underdetermined. In fact, \textit{meaning is always underdetermined} \cite{Quine:WordObject}. Because objects are externally determined, communicators are always referring to any number of potential social or material objects. Meaning happens when relations between these social or material objects correspond to relations between experiential correlates (of these social or material objects) for both communicators. This correspondence is precisely what inscription must then produce. We claim, then, that \textit{in a vacuum (with no other social or material forces), inscription by a meaning-agent produces social or material objects with determinations that correspond to the determinations of experience for that meaning-agent}. This is our ``postulate of inscription.''

The material objects created by inscription are totally concrete (in contrast to their experiential correlates). All possible properties (e.g. location, time, composition) are determined. Some of these determinations happen arbitrarily because they have no correlate in the mind. For example, the chemical makeup of the paint in a painting is usually not determined for the painter.

This gives us \textit{a genealogy of the object}: the object begins in the experiential world as totally abstract, an undetermined `thisness'. Over time, it accrues determinations, becoming gradually more concrete. Eventually, it proceeds out into the social-material world, where it is at its most concrete. This procession redetermines the structures of the social-material world, creating a cyclic effect whereby inscription by agents conditions future concretization.

\subsection{The social object}\label{sec:theory:social}
This theory of meaning also gives us the appropriate tools to describe the object as it passes beyond the experiential world and into the social (material) world -- in other words, the object as a \textbf{social object}. This discussion of the social object will prefigure our forthcoming analysis of values and categories insofar as they are themselves social objects -- most pertinently, for example, morality.

Recall our ``postulate of inscription'': inscription reproduces the determinations of the object it inscribes.
Now, consider a material object. It has a variety of determinations -- color, location, composition, and so on. Only some of these are ever experienced by individuals. The sum of these experiences determines a \textit{social} object. In general, a social object is a sum of determinations over individuals in a society. It is reasonable to ask in what sense the social object `really' exists. Who is the social object an object for? We claim that it is useful to posit an abstract model of society as a meaning-agent. We can treat this abstraction much like we treat individuals. It has its own concepts, signs, and objects. We do not have to be ontologically committed to its existence for it to be useful. Generally speaking, this abstraction is something like our collective ground of meaning as a society.

We give this abstraction the name \textbf{social totality}, emphasizing its nature as a system of meaning. Then, \textit{the objects of the social totality are the social objects.} \textit{Moreover, the }\textbf{concepts}\textit{ of the social totality are many of the }\textbf{contexts}\textit{ about which we have been writing.} For example, the language-game, which operates as as \textit{context} for an individual, is a \textit{concept} for the social totality \citep{Gadamer:TruthAndMethod, Heidegger:BeingAndTime}. \textit{In fact, all social contexts are concepts of the social totality.} This means that the social totality accrues a variety of concepts in which to determine objects.


Because of this diversity of concepts, when considered as a meaning-agent, the meanings of the social totality are highly indeterminate. Thus, the social totality is perpetually in conflict with itself. This constant struggle is the social history of the object~\citep{foucault2002order}. In a certain sense, then, the truth of social totality is eternal self-contradiction~\citep{Hegel:PhG}. But it is a powerful kind of self-contradiction comprised by multitudes of information grouped according to their contexts. So we must be precise: to understand the contradiction involved in a concept, we must understand the histories, peoples, and diverse points of view which go into making a concept what it is. 

Let us now consider an example: how does a child learn what `fairness' means? `Fairness' begins as a social structure -- a \textit{social object} for the \textit{social totality}. A child learns certain things from their society about norms of behavior.
Slowly, they develop a \textit{concept} of effort as ``fairly'' \textit{signifying}' reward \citep{Rawls:TheoryJustice}. These concepts -- norms, deservingness, etc. -- amalgamate around the word `fairness' as `fairness' is \textit{concretized}. When this child says, ``That's unfair!'', they are doing more than pointing out some behavior that violates the norms. They also invoke all concepts with which `fairness' is related. Finally, the way this child writes and acts will reflect the myriad ways they perceive ``fairness'' as an aggregate of many experiences. The potential misconceptions or priorities this child has about fairness will go back into the world -- they will be \textit{inscribed}.




\section{The Meaning Model}\label{sec:model}


The LLM is both a social-material object and a meaning-agent.
It is natural to ask why a model can be considered a meaning-agent -- we do not explore this here, but instead in $\S$\ref{appendix:possibility}.
Humans inscribe the LLM as a social-material object. In this process, the formal structure of the statistical intelligibility in language \citep{doi:10.1080/00437956.1954.11659520, Lonergan:Insight} is inscribed into a statistical machine -- the LLM. At an abstract level, this is very similar to how language itself captures the intelligibility of the world. Other similar examples form cornerstone human cultural technologies \citep{YiuGopnik:CulturalTechnologies}. In this sense, a latent LLM is an object for humans. But an LLM, once activated, becomes something more: a meaning-agent \cite{Arcas:MachinesBehave}. We will proceed to explore this process. In so doing, we go from the model as a scientific object to apprehending the model in experience, reading the objectivity of the model as textual \citep{Hegel:PhG, Gadamer:TruthAndMethod, Heidegger:BeingAndTime}. 


\begin{figure}
     \centering
     \begin{subfigure}{0.45\linewidth}
         \centering
         \includegraphics[height=7cm]{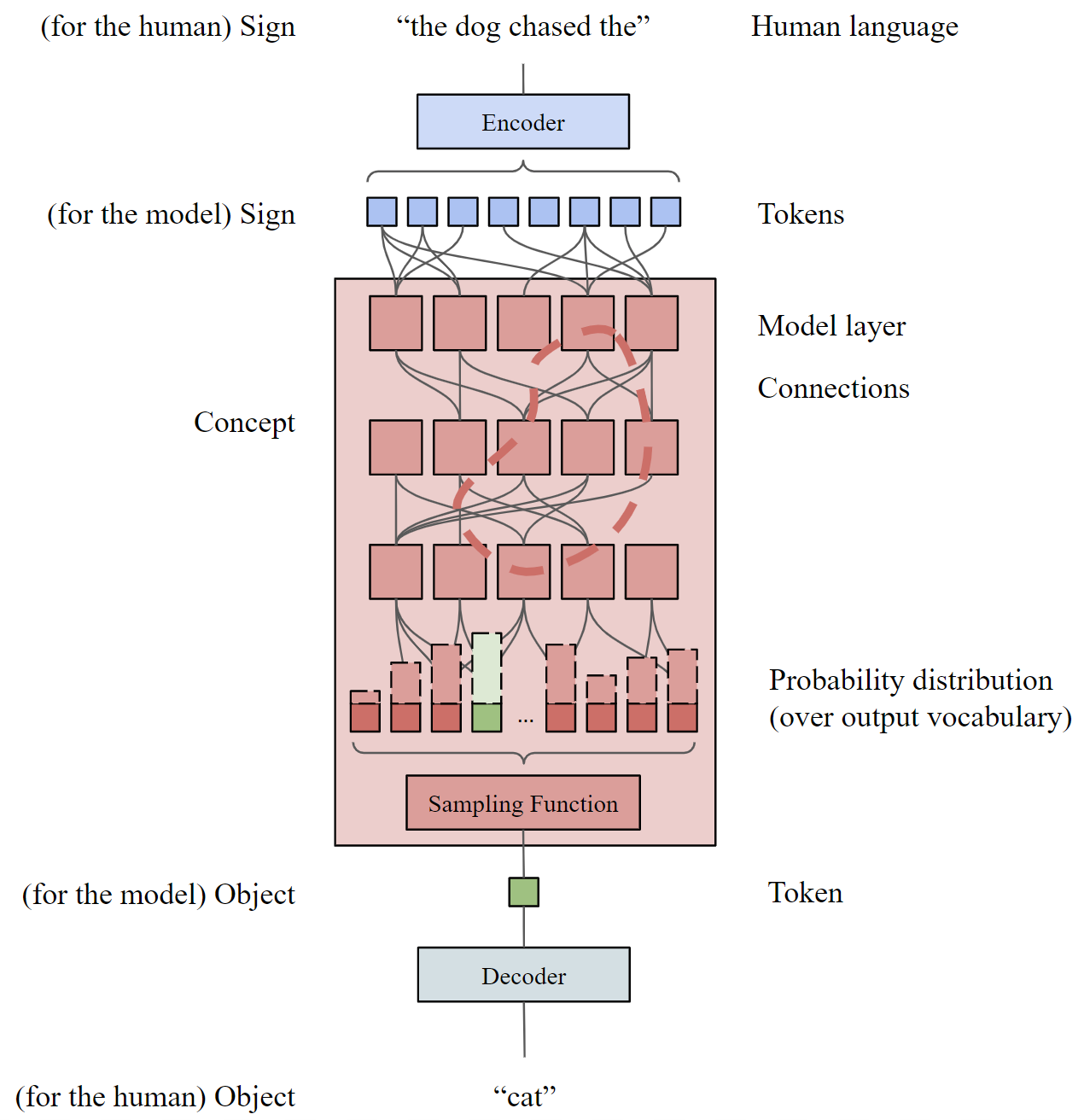}         \caption{\raggedright{\textit{Inscription.} A forward-pass mediates the picking-out of an object from a sign.}}
         \label{fig:inscription}
     \end{subfigure}
     \hspace{2mm}
     \begin{subfigure}{0.45\linewidth}
         \centering
         \includegraphics[height=7cm]{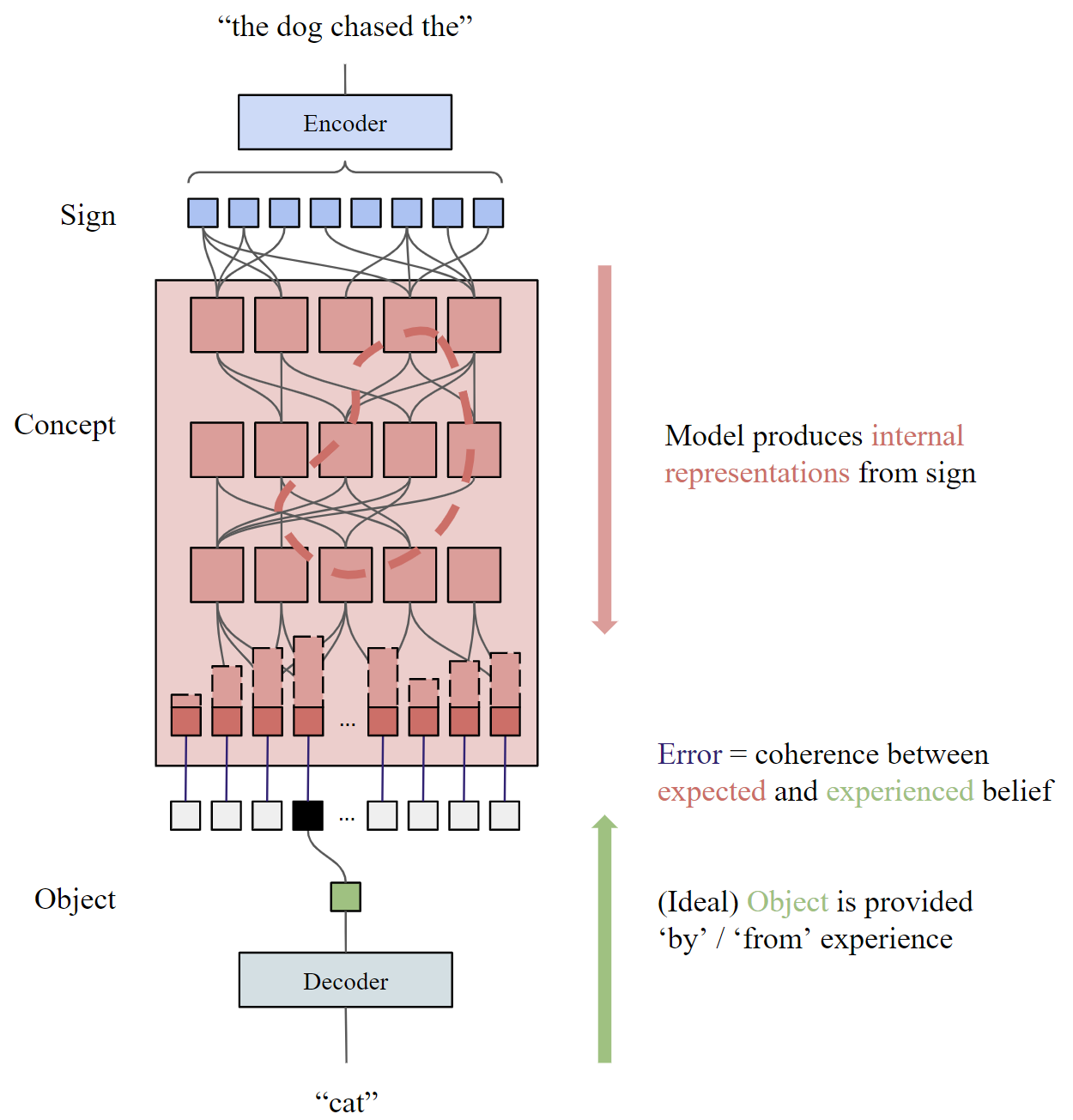}
         \caption{\textit{Concretization.} Backpropagation adjusts model concepts in response to error.}
         \label{fig:reification}
     \end{subfigure}
     
    \caption{Inscription and concretization represented in LLMs. See $\S$\ref{sec:model} for a more thorough exploration.}
    \label{fig:conc_inscrip}
\end{figure}

An LLM $f$ is a function from a variable-length sequence of tokens $\vec x = \{ x_1, ..., x_n \}$ to a single token $\hat{y}$.\footnote{Although this may be for next-token prediction, it also applies to schemes like self-supervised masked token prediction which do not observe a strictly sequential relationship.}
By this definition, the sampling procedure which selects a particular token from a probabilistic distribution over the output vocabulary is part of the LLM; different sampling procedures produce different output tokens.
\textbf{The LLM is a repository for concepts which are activated by the sign $\vec x$ to pick out the object $\hat{y}$. }
The sign and object are of the same substance; they are all tokens in the model's ``phenomenal world'', such that the model may encounter an object as a sign.
Autoregressive generation, for instance, is a series of operations $\{x_1, ..., x_p\} \to_f \hat{y}_1, \{x_1, ..., x_p, \hat{y}_1 \} \to_f \hat{y}_2, ... \{x_1, ..., x_p, ..., \hat{y}_{n-1}\} \to_f \hat{y}_n$.
In this case, the model encounters the object of its world which it has previously picked out as part of the sign in another operation.
Empirical work in model interpretability demonstrates that LLMs and neural networks of sufficient scale generally develop concentrations of features developed from statistical co-occurrences throughout training~\citep{Frankle2018TheLT,Kauffmann2019FromCT,Zhang2020ASO}. These concentrations represent the models' concepts.

In LLMs, \textit{concretization} corresponds to weight update during training~(\ref{fig:reification}). At each timestep in training, the following steps are observed.
First, a pair $\langle \vec x, y \rangle$ is retrieved from the training dataset, where $y$ is the ``ground truth'' token.
Second, the model performs a \textit{partial} forward pass to obtain the probability distribution over the output vocabulary $\vec{p}$. 
Third, the difference between $\vec{p}$ and $y$ is computed as the error (encoding $y$ to make the computation feasible). 
Fourth, this error is propagated back throughout the model such that on another such partial forward pass the error would have been smaller.
During training, the model is repeatedly presented with co-occurrences of sign and object. Accordingly, the model develops internal clusters of representations (\textit{concepts}), thereby undergoing \textit{concretization}.
On the other hand, \textit{inscription} corresponds to inference $\vec x \to_f \hat{y}$, in which the model writes back `into the world' such that it may potentially encounter it again (for instance, in autoregressive generation)~(\ref{fig:inscription}).
This strict separation between concretization and inscription is in stark contrast to their intertwinedness for humans.
However, the observed phenomenon that LLM-generated content is increasingly infiltrating datasets used to train LLMs~\citep{Shumailov:Curse} is an instance of inscription and concretization simultaneously at work.






\section{The Moral Model}\label{sec:moral}

We see LLMs as microcosms of the social totality. This means that LLMs retain the complex history and diversity of information used in determining any social object.
As a result, \textbf{unaligned LLMs have already grasped morality in concept}, provided that they faithfully represent the social totality. In practice, there is often an intelligible difference between LLMs and the social totality: LLMs may not be trained on representative data or with large enough quantities of data, and may fail to adequately reflect the data.
However, once we accept the basic premise that LLMs resemble the social totality, we understand that we are not aligning `amoral' models with `human' values. Rather, alignment chisels away existing moral systems within unaligned models, leaving behind the
idiosyncratic values
of the aligners.
We must therefore make normative arguments as to which values we should align towards. On the other hand, reading LLMs as microcosms of the social totality gives us another account of the dangers latent in LLMs. The social totality is unfair and unjust,
predicated on a history of oppressive relations.
The inaccuracy of LLMs with respect to minority groups is a sharpening of already existing power structures. This becomes especially salient when understanding small communities lacking social objects determinate in text. The social totality may possess a statistical intelligibility; but the statistical intelligibility congealed within the LLM is really the intelligibility of the social totality determinate in text. As a practical example, cultures with strong oral traditions that are not recorded in the digital datasets on which LLMs are often trained are lost among the many other voices in the history of a social object.
This is uncomfortable, and certainly grounds for alignment.

    


\subsection{Towards a genealogy of morals}
The historical development of AI ethics has largely been guided by an intuitive understanding of morality. This intuition is prey to a litany of biases. To combat this, we must develop a rigorous grounding for ethics through the lens of meaning.

In $\S$ \ref{sec:theory} we described social objects as consequences of diverse viewpoints and histories.
\textit{Values are social objects} -- objects for the social totality (see $\S$\ref{sec:theory:social}) with a complex genealogy.
Every system of values is also a social object, even such axiomatic systems as deontologies \citep{Kant:CritiquePureReason}. As social objects, both values and value-systems are always being concretized and inscribed within the social totality. Values and value-systems exist only insofar as they have a social history \citep{Nietzsche:GenealogyMorals,Scheler:Ressentiment}.
Thus, in practice, we are always renegotiating what values are.
Like all other objects, values are externally determined within contexts.
Thus, understanding a value means understanding the diverse histories which have dynamically constituted it within the social totality.

This means that \textbf{a unitary conception of value is unattainable.}\footnote{See $\S$\ref{appendix:possibility} for an exploration of the notion of unitarity.}
We must confront the impossibility of this maxim head-on if we ever wish to build truly ethical systems.
Value pluralism is one way of confronting this impossibility. Because it already captures this metaethical maxim \citep{Berlin:FourEssaysLiberty, Rawls:TheoryJustice}, value pluralism has proved useful for the project of AI alignment \citep{Sorensen:ValueKaleidoscope,Marchese_2022}. 
Building on this, we distinguish between two views on value pluralism which attempt to describe what AI alignment has accomplished in moral philosophy: pluralism \textit{in content} and pluralism \textit{in concept}.

\paragraph{Pluralism in content} attempts to construct contents for values; it aims to create a \textit{non-unitary} set of \textit{unitary values}.
This is the paradigm of current AI alignment: unitary values either are or are not embedded into models.
In theory, pluralism in content should create ethical systems which, among other things, promote social, political, and value pluralisms.
In this attempt to construct contents for values (which is, under our schema of values as social objects, theoretically impossible), what all-too-often happens is the importation of certain partial contexts into values.
In practice, this means that pluralism in content often encourages sets of unitary values that privilege the status quo and fail to challenge institutional systems where necessary.
RLHF~\citep{Bai2022TrainingAH} is a good example of this tendency in alignment/fairness; there already exist studies, for instance, showing that systems aligned via RLHF absorb new political biases \citep{Casper:OpenProblemsRLHF} from human feedback.
One can only expect such externalities when manipulating/aligning models on the basis of content, which is necessarily particular.
See $\S$\ref{appendix:westernmorality} for a discussion on how RLHF can be interpreted as forcing an erasure of the social history of values.

\paragraph{Pluralism in concept} attempts to replicate the external determinations of values; it aims to create a \textit{non-unitary} set of \textit{non-unitary values}.
The crucial complication here is that a value such as fairness is no longer treated as unitary. Instead, it is regarded as a social object with a diverse history.
We ourselves have idiosyncratic and unitary values. Insofar as they are idiosyncratic, they differ from person to person. Moreover, insofar as value pluralism is commonly held in AI ethics, we should be motivated to create ethical systems that model the diversity of human thought on values \citep{Sorensen:ValueKaleidoscope}. As it happens, modern unaligned LLMs are already close to being such systems. Now, some alignment is necessary for many practical purposes.
In this regard, a system like Constitutional AI \citep{Bai:ConstitutionalAI} which ``queries'' the LLM for a specific set of moral constitutional principles (and ``self-improves'', or more accurately, morally narrows in content) is closer towards a pluralism in concept. Importantly, this ``constitution'' could take any form, because the LLM already understands these moral principles \textit{in concept}.



\subsection{LLMs grasp morality in concept}
In $\S$ \ref{sec:model}, we discussed the model as a meaning-agent. We now proceed to argue that the ground of experience of the model makes it broadly representative of the social totality.
To begin, we must recognize two points. First, the social totality is not a material totality of circumstances (it is not the state of affairs).
It is the collection of social objects. 
Second, by virtue of this smaller sphere, the social totality has a statistical intelligibility practically learnable by statistical machines (LLMs).

Considering just social inscriptions (i.e. communication in general), the external determinations between inscription of a communicator and concretization of a communicated-to are mediated by the causal capacity of agents in prioritizing contexts.
As a statistical law, this causal capacity will tend towards prioritizing shared experience. As a result, the objects of the social totality are woefully immaterial, or at best woefully ambiguous
in any particular context. This is, however, helpful to us: it means that if we are interested in understanding the social totality, disembodiment will do.

We could choose to make a given model more embodied and likely more effective on certain tasks as a result. However, this embodiment is particular; it bucks against shared experience. This is in fact our curse as human observers -- we are embodied, all too embodied. Thus, not only is disembodiment good enough for a statistical machine built to understand the social totality -- it is almost required.

The social totality can take quite a concrete form.
It is not too difficult to imagine the corpus of all text as being, if not co-identical with, at least a very good model of the social totality. The corpus of all text is home to history, diversity, contextual determinations, abstract social objects, and so on. Most importantly, it also exhibits a more tractable statistical intelligibility \citep{Lonergan:Insight}. Insofar as a modern LLM approximates the corpus of all text, it acts as a concrete oracle of the social totality.

The nature of certain social objects is deeply textual. As perhaps the clearest example, gender is a social object for which we privilege language as the mode of truth. Attempting to `read' a person's gender from visual cues or social behaviors many now accept to be immoral (or at the very least harm-causing) inasmuch as it is the root of all misgendering.
Text is the root of truth for gender.
And in fact, we see that LLMs have begun to develop limited understandings of gender. For example, a translation of the neuter pronoun to languages without one now might demonstrate the ambiguity in gender (i.e. his/her) \citep{Arcas:MachinesBehave}. This textuality of gender gives us a theoretical basis to understand, for instance, why automatic gender recognition (AGR) is harmful \citep{Keyes:MisgenderingMachinesAGR}. AGR attempts to redetermine a concept in image whose truth `should' be determinate in language.

This passage between the normative and positive is much more general -- a (false) positive claim about a social object can easily become a normative transgression. The story is similar with values. Even though there is no unitary narrative that tells us what good is, we privilege language as the mode of truth above others. Many horrible acts throughout history have been committed for `good' (unobjectionable) values; we are better guided by our idealistic ideas \textit{in text} of what value should be.

In this sense, not only do LLMs \textit{approximate} a concrete oracle of the social totality by representing
the totality of text; LLMs \textit{are} concrete oracles for \textit{concepts in the social totality that have a truth determinate in language}. Less abstractly, \textbf{LLMs grasp morality in concept. }Language is the privileged domain of signification for many social objects \cite{harnad1990symbol}. These determinations in language give LLMs a cutting insight into the nature of social objects like values and categories, an insight which should not be carelessly thrown aside for our own constructed contents and desires.





    

\section{Conclusion}\label{sec:conclusion}

Like humans, LLMs are meaning-agents. We must understand the way in which LLMs mean through the lens of a general, agent-agnostic, multimodal theory of meaning. We offer one solution, a model of signs, concepts, and objects in the world of experiences that evolve over time. For us, objects in the social-material world are the consequences of long histories. Individuals immediately grasp objects as merely abstract, pure `thisness', then concretize their experience of objects over time. When these objects are inscribed into the social-material world, their social correlates -- the social objects -- change; what fairness is changes as meaning-agents inscribe new determinations into its social existence. We make one essential claim: contrary to our intuitive understanding of meaning, we are not referring to the same `real' objects when we communicate. Rather, we are indicating structures of possible experience. These structures must enter the `real' world in a way roughly corresponding to their experiential correlates (this is our ``postulate of inscription'') for communication to occur.

Our general theory of meaning can ground future work in AI ethics, fairness/bias research, semiotics, and philosophy. In alignment, we should continue to explore AI feedback, along the lines of Constitutional AI. In AI ethics, we should begin to consider the liberatory possibilities of models that already understand us in concept: gender-affirming content, disruption of hegemonic structures, historicization of values, etc.
In moral philosophy, we can work towards a theory of values as individual or collective projects informed by their existence as social objects.
Finally, there is an epistemological need our theory makes apparent: we must work towards a better understanding of how models and humans put their meanings out into the world; how inscription happens.

Thus LLMs grasp and reproduce the values of a statistically intelligible society. LLMs have a powerful access to the concepts of our values and categories. LLMs `embody' the social totality; the way LLMs mean is effectively the same way societies mean. Like societies, LLMs are comprised by a diversity of contradictions stemming from the complex genealogy of social objects. We therefore say that \textbf{LLMs grasp morality in concept.} Unaligned LLMs thus already can reflect a myriad of value-systems, and can serve as potential objects of study to descriptively understand what human values are. When we proscriptively inscribe our idiosyncratic values into LLMs, we make LLMs ``victims of meaning.'' We should be clear and explicit about when we do this and why. Ultimately, we must better leverage the capabilities of LLMs as modern oracles of the social totality.




\appendix
\section{Supplementary Material}

\subsection{The possibility of non-intentional and non-anthropocentric meaning}\label{appendix:possibility}
Despite the generality of our theory of meaning, many theories of meaning require, implicitly or explicitly, some sort of ``consciousness'', ``mental states'', or ``intentionality''. 
There is much disagreement over whether contemporary LLMs can be said to have any of these, but we contend that our understanding of LLM meaning is significant regardless of one's views on these questions.
Through our theory, we can pivot away from the question ``Do models have mental states / agency?'' and instead ask ``Do models have sign-object relations?''
What enables us to make this change?
Beyond the mere tautology of defining meaning in an agent-agnostic way, we must justify heuristically why this definition of meaning meets our expectations if it admits nonconscious egos.

We begin by laying the groundwork with intentionality. Intentionality -- the directedness of minds towards mental objects ~\citep{Brentano:PsychologyEmpiricalStandpoint} -- has played a central role in previous theories of meaning ~\citep{Dreyfus:Intentionality}. The most influential of these theories of meaning in modern AI~\citep{Andreas:LMsAMs, Bennett:ComputationMeaning}) is the Gricean theory~\citep{Grice:Meaning}, in which meaning consists in the idea an utterer intends a hearer to recognize. It is presaged by the Husserlian phenomenological theory~\citep{Husserl:CartesianMeditations}, which similarly understands utterers as modelling the minds of hearers. Intentionality manifests itself through act-term pairs (noesis-noema, cogitationes-cogitata). Terms stand for objects.\footnote{This Husserlian structure of intentionality has an obvious parallel with our theory of meaning: concepts are acts, terms are signs, objects are objects, and the standing for between terms and objects is signification.}

Importantly, intentional acts grasp their terms (and accordingly, their objects) unitarily. Unitarity has been a central theme throughout our paper, so we should provide a definition (largely a small refinement of the colloquial definition) here before proceeding: something `unitary', in our usage, is essentially something `coherent' `cogent', or `harmonious'~\citep{Husserl:CartesianMeditations}. It has a direct intelligibility \citep{Lonergan:Insight}; there is something which can be explained. The unitarity of individual concepts is a direct result of the unitary grasping of terms by intentional acts.

Now here is the question we must answer: is unitarity necessary for an utterer to simulate the recognition of an idea by a hearer? We argue that the answer is \textit{no}, as long as the conception of the hearer is non-unitary. In practice, this means that a meaning-agent can contradictorily model a hearer as potentially recognizing many different meanings for the same term. Empirically speaking, we know that language models simulate agents. In many contexts, they simulate agents fairly well, and `act' in ways that can appear relatively coherent \citep{Andreas:LMsAMs}, in spite of their general non-unitarity.

Moreover, insofar as we are concerned with how LLMs come to know social meanings as whole, unitarity becomes even less necessary. An analysis of social meanings (see $\S$ \ref{sec:theory:social}) quickly shows their diversity makes them scattered and non-unitary. Paradoxically, a single unitary agent could not possibly hope to understand the social totality. In this way, the unitarity of intentional agents is actually limiting. We can compare an LLM acting in this way to a nondeterministic machine, exploring all possible meanings at once, sprouting and culling paths towards meaning all the while.

In $\S$\ref{sec:theory:diachronics}, we discussed the prioritization of contexts  as a method to disambiguate between potential meanings of a sign. This disambiguation is really unitarity; unitarity within a context. What this means for models is that LLMs fail to prioritize contexts (on one account, this is because they have no Umwelt -- a material world of important factors for a subject ~\citep{Uexkull:Umwelt})). This condition is equivalent to their nonunitarity. Thus, we can situate prioritization as made possible entirely by intentionality -- intentionality is directedness and focus. Where this directedness of intentionality fails -- e.g. in the case of mental illness~\citep{Deleuze:ThousandPlateaus} -- nonunitarity results.

So, \textbf{models can mean} -- they are meaning-agents. They are not intentional meaning-agents: like the social totality, they lack causal capacities, Umwelten, and prioritization. There is one eternal objection: whatever meaning we ascribe to LLMs under any system, however clever, is actually ascribed by us. When we say that LLMs mean X, it is actually that we mean that LLMs mean X. To this objection, we make the even more eternal objection that the same is true of humans. To deny that humans mean because we ascribe meaning to them is to plunge into solipsism, and in avoiding this we have no choice but to also admit the LLM.

As a final note, taking a heuristic, functionalist look at what meaning is, meaning appears in playing the language-game. This notion is captured, for example, by the Turing test, which it is quite uncontroversial to say modern LLMs pass (from as early as 2014) \citep{Warwick:MachinesThink}. We can go even further and describe AI as having human-like properties for reflection of interlocutors \citep{Sejnowski:ReverseTuringTest}.

We have played games with machines for as long as we have machines capable of so doing -- ultimately, the language-game may prove to be no different.

\subsection{Discussion: Western / Christian Morality}
\label{appendix:westernmorality}
AI alignment may take as its goal to imbue models with certain abstract moral principles.
Some seemingly relatively uncontroversial candidate principles include fairness, kindness/charitability, equality.
But they are precisely uncontroversial because of their historical development through the bourgeois revolutions of the early nineteenth century.
Their purported ``common-sense'' ubiquity in Western nations and cultures is, in one view, a result of the French Revolution~\citep{Scheler:Ressentiment}; the conservative forces which opposed fairness and equality were silenced by bloodshed. 
In another view, this ubiquity is a result of an inversion of the values of Greco-Roman society by those who it dominated -- ``slave revolt in morals''~\citep{Nietzsche:GenealogyMorals}.
The general point is that these principles have a social genealogy involving resistance and contradiction, and that the artifacts of this genealogy are embedded within the social totality that LLMs congeal.
Certainly, one could ask an LLM, ``What is Nietzsche's argument for the geneology of fairness as a value?'' and receive a valid response \textit{in content}. 
For instance, ChatGPT writes:
``Nietzsche contends that fairness, or the concept of justice, originated as a reaction to power dynamics between dominant and oppressed social groups. Fairness evolved from a means of asserting the worth of the oppressed into a cherished moral virtue, resulting in a transvaluation of values where the weak's emphasis on fairness came to dominate society.''
Yet, in a more radical sense, the \textit{concept structure} of the LLM -- i.e. the statistical arrangement of different data sampled from across the social totality -- stands for or represents the genealogy of fairness, or indeed any moral value.
In this sense, aligning a model with RLHF might amount to \textbf{forcing an ``uncritical'' acceptance of the value, or a structural ``forgetting'' of the value's genealogy}.

\bibliography{bibliography}
\bibliographystyle{apalike}





\end{document}